# Deep Adaptive Network: An Efficient Deep Neural Network with Sparse Binary Connections

Xichuan Zhou*, Shengli Li, Kai Qin, Kunping Li, Fang Tang, Shengdong Hu, Shujun Liu, Zhi Lin

***Abstract*—Deep neural networks are state-of-the-art models for understanding the content of images, video and raw input data. However, implementing a deep neural network in embedded systems is a challenging task, because a typical deep neural network, such as a Deep Belief Network using 128×128 images as input, could exhaust Giga bytes of memory and result in bandwidth and computing bottleneck. To address this challenge, this paper presents a hardware-oriented deep learning algorithm, named as the Deep Adaptive Network, which attempts to exploit the sparsity in the neural connections. The proposed method adaptively reduces the weights associated with negligible features to zero, leading to sparse feedforward network architecture. Furthermore, since the small proportion of important weights are significantly larger than zero, they can be robustly thresholded and represented using single-bit integers (-1 and +1), leading to implementations of deep neural networks with sparse and binary connections. Our experiments showed that, for the application of recognizing MNIST handwritten digits, the features extracted by a two-layer Deep Adaptive Network with about 25% reserved important connections achieved 97.2% classification accuracy, which was almost the same with the standard Deep Belief Network (97.3%). Furthermore, for efficient hardware implementations, the sparse-and-binary-weighted deep neural network could save about 99.3% memory and 99.9% computation units without significant loss of classification accuracy for pattern recognition applications.**

## I. Introduction

THE approach of Deep Belief Networks (DBN) demonstrated its remarkable performance for feature extraction and pattern recognition in the last a few years [1-4]. The fundamental computations of a feedforward Deep Belief Network were a large number of floating-point multiplications between the matrix of weight parameters and the input data, which can be implemented using clusters of central processing units (CPUs) or general purpose graphics processing units (GPUs) in powerful computers and cloud servers [5]. However, for embedded computer vision and cognitive applications, where memory and power budget is limited, a more efficient network architecture and training algorithm would be ideal, one that is designed for efficient hardware computing, and requires only fix-point computations and much less number of parameters, allowing larger networks to be implemented using embedded systems for cognitive applications.

For the last a few years, researchers invented different digital hardware to accelerate feedforward neural computations for real-time feature extraction and pattern recognition [6-13]. Himavathi and Byungik presented different digital implementations of the feedforward neural networks using reconfigurable field-programmable gate arrays (FPGAs) [6-8]. Sanni presented an FPGA-based implementation of a Deep Belief Network for character recognition using stochastic computation [8]. To further improve the efficiency and processing power, Morgan, Jonghong and Chen proposed respective digital architectures and implementations using application specific integrated circuits (ASIC) [10-13].

Recently, Merolla and his colleagues in the IBM company took advantage of cutting-edge integrated-circuit technology and built a large-scale neuromorphic chip, known as the TrueNorth chip, for feature extraction and pattern recognition applications [14]. The TrueNorth chip is a breakthrough which integrates a million neurons and 256 million synapses in a single ASIC; however, computer vision researchers indicated that the chip had limitations for high-dimensional computer vision and pattern recognition applications [15]. The TrueNorth chip adopted a simple quantization strategy to represent the weight parameters of a classic Deep Belief Network with two-bit integers, which saved memory footprint but resulted in notable loss of classification accuracy [16].

Through, the research of custom designed hardware for deep neural networks is booming, as far as we know, few attempts have been made at the algorithmic level to optimize the training and learning process of network parameters for efficient hardware implementation [17]. To efficiently implement a feedforward neural network, the key algorithmic consideration was the number of neural connections, which equals the number of weight parameters and the number of multiplication operations. The Deep Belief Network has fully connected neurons between adjacent layers (Fig. 1), resulting in approximately $O(n^2)$ memory and computational complexity [7]. To relieve the memory and computational bottleneck, we exploited the sparsity in the neural connections. By adopting the usual assumption that some features extracted by the deep neural networks are negligible for classification and pattern recognition, our approach, named as the Deep Adaptive

X. Zhou*, S. Li, K. Li, F. Tang, S. Hu, Z. Lin, S. Liu are with the Chongqing Engineering Laboratory of High Performance Integrated Circuits, College of Communications Engineering Chongqing University, Chongqing, China, 400044.

K. Qin is with the Division of Remote Sensing Technology Application, CNNC Beijing Research Institute of Uranium Geology, Beijing, China.

*Asterisk indicates corresponding author*, e-mail: xichuan.zhou@qq.com.

This research was supported in part by the national natural science foundation of China (contract 61103212, 61471073), and by the post-doctoral science foundation (contract 2012M521678, 2013T60836 and XM20120036).

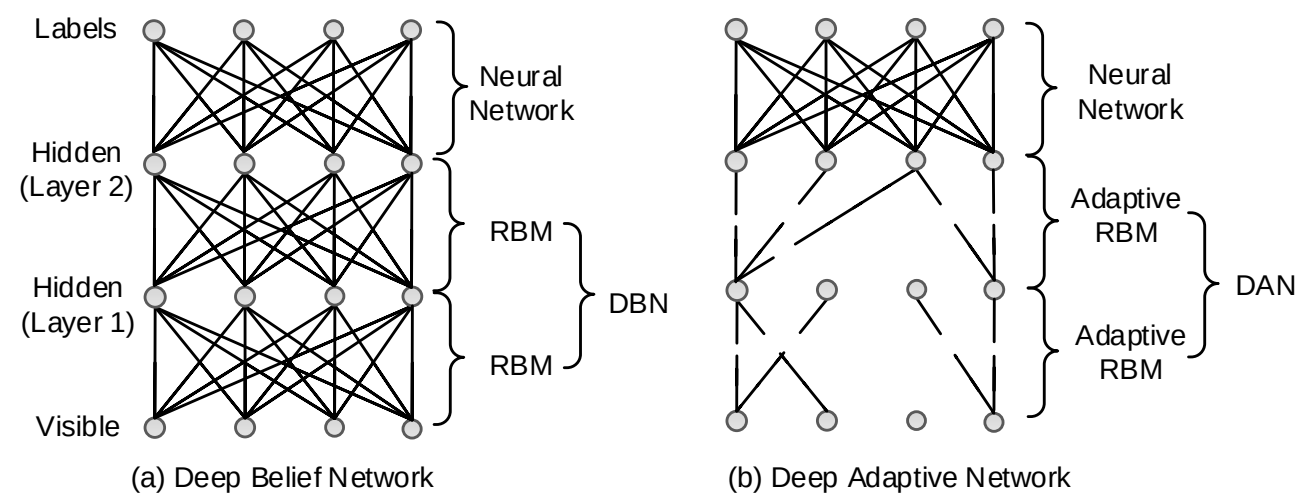

Fig. 1 Deep neural networks implemented for cognitive applications. The Deep Belief Network (a) is full-connected between adjacent layers, whereas in a Deep Adaptive Network (b), the connections with zero weights are removed, leading to sparser architecture, which is efficient for hardware implementation. A layer of classic neural network is concatenated with the deep neural networks for classification.

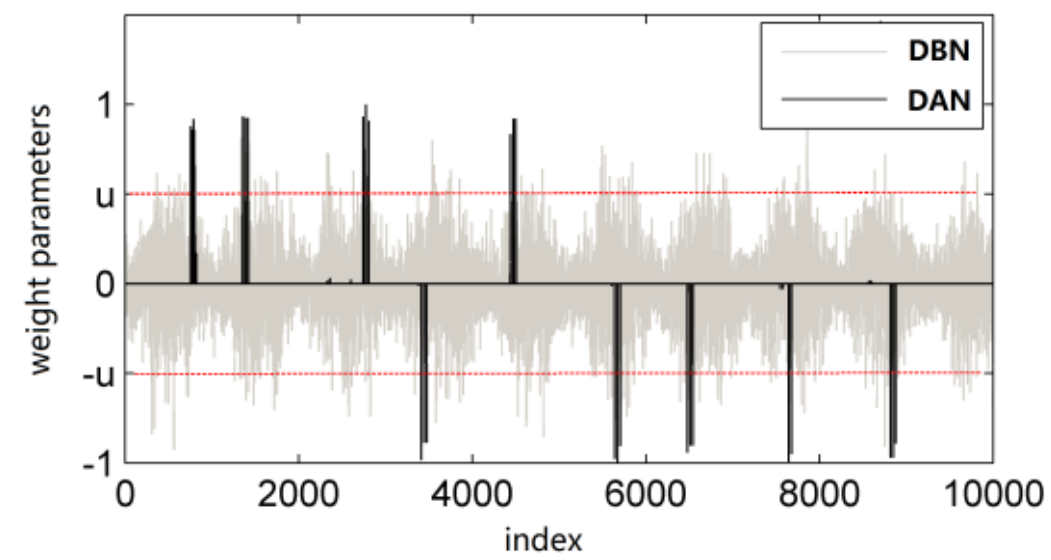

Fig. 2. The weight parameters of a DBN and a DAN trained with images of MNIST handwritten digits. The 784 × 784 weight matrix were lined up in a column by column fashion. The values of DAN weights are sparse and concentrated on a few important neural connections, leading to robust sparse and binary representation of the neural connections.

Network (DAN), adaptively reduces the weight parameters associated with negligible features to zero, which can be omitted for efficient hardware implementation.

The second algorithmic consideration for efficient hardware implementation is data representation, which is an essential tradeoff between accuracy and cost. Digital implementations of the feedforward neural networks usually use fix-point representations. Researches indicated that, deep neural networks trained with limited-precision could suffer from significant loss of accuracy [18-20]. Gupta indicated that, in most cases, 8-bit width was necessary for training a deep neural network to achieve convergence and adequate accuracy [20]; however, since the sparse weights of the proposed Deep Adaptive Network are naturally separated in to roughly three groups, i.e. (1) close-to-zero weights (2) large positive weights and (3) negative weights with large absolute values (Fig. 2), one can robustly threshold and represent the weight parameters using single-bit integers without notably influence on pattern recognition accuracy.

The proposed method could potentially reduce the memory and computational complexity of the feedforward neural network by about 99%. Our main contributions are as follows:

1) We introduce the Deep Adaptive Network, a method to train a deep neural network with sparse connections by incorporating a novel mixed-norm weight-decay regularization process. The DAN algorithm adaptively reduce the negligible weight parameters to zero, which can be omitted for efficient implementations.
2) The remained important weight parameters can be robustly thresholded and represented as binary integers without significantly loss of classification accuracy. The binary weights can significantly reduce the memory complexity and replace complicated floating-point multiplications by simple logic operations.
3) The Deep Adaptive Network can extract sparse features, because it is designed to reduce all the weight parameters associated with unimportant features. The negligible features become close to constants, which is efficient for hardware implementation.

The rest of this paper is organized as follows. Section 2 introduces the related work. Section 3 presents the formulation and algorithm to estimate the sparse-binary weights and features. Section 4 presents the experiments of recognizing handwritten digits in the MNIST dataset. We conclude the paper in Section 5.

## II. Related work

Early attempts of incorporating sparsity in neural networks were inspired by the researches in visual cortex, where the neural activity was found to be sparse [21]. For learning algorithms of sparse deep neural networks, Ranzato proposed am encoder-decoder architecture to learn sparse representations [22]; Recently, Lee developed a variant of the Deep Belief Network to learn sparse representations of input images, and found that the selected sparse features had some properties similar to visual area V2 [23]. Ji proposed a sparse-response DBN based on the rate-distortion theory, which attempted to encode the original data using as few bits as possible [24]. Generally speaking, these researches focused on exploiting the sparsity in the extracted features (or activated neurons); however, motivated by learning efficient architectures of a deep neural network for embedded implementations, our work focused on exploiting the sparsity in neural connections.

Binary weights would bring great benefits to neural network implementations by replacing the huge number of multiply-accumulate operations by simple accumulations, as floating-point multipliers are the most space and power-hungry components of the digital neural network implementations [25]. Therefore, training deep neural networks for binary weights has been the subject of very recent works. Daniel and Courbariaux trained the deep neural network with backpropagation and expectation backpropagation respectively [17, 26, 27]. By comparison, instead of training binary weights, our method aims to learn sparse weights and reduce the small weights to zero. As a result, the weight parameters were adaptively separated into three groups, which can be robustly thresholded and ternarized as (-1, 0, 1). With most zero weights omitted, the small proportion of important weight parameters can be represented using single-bit integers.

## III. Methods

The computation of the deep neural networks for pattern recognition consists of two stages, i.e. the training stage and the evaluating stage. The training algorithms usually approximate the parameters offline using high performance computers. At the evaluating stage, a multi-layer feedforward neural network is implemented based on the approximated parameters for real-time feature extraction. Many hardware accelerators have been recently proposed to implement the feedforward neural network;

however, few researches focused on the training algorithm for efficient hardware implementations. In this section, we propose the DAN training algorithm which addresses the challenge of memory and computation bottleneck for building a feedforward neural network. To explain the DAN formulation and algorithm, we first briefly introduce the original DBN algorithm.

### *A. Deep Belief Network and Restricted Boltzmann Machines*

The DBN is a state of art training algorithm of the deep neural networks widely used for feature extraction. A DBN is constructed by stacking multiple layers of Restricted Boltzmann Machines (RBMs) and using the output (hidden layer) of the previous RBM as the input (visible layer) of the next RBM (Fig. 1). Higher layers tend to encode more abstract features, which are typically informative for classification tasks.

A standard RBM consists of two layers of binary units, a matrix $\mathbf{W} \in R^{n\times d}$ is defined as connection weights, where $w_{i,j}$ represents the connection between visible unit $v_i$ and hidden unit $h_j$. The parameters $b_j$ and $c_i$ are the bias for the visible and hidden unit respectively. Given the vector forms of hidden units as **h**, visible units as **v**, bias as **b** and **c**, the energy of a configuration (**v, h**) can be written as

$$E(\mathbf{v},\mathbf{h}) = -\mathbf{b}^{\mathrm{T}}\mathbf{h} - \mathbf{c}^{\mathrm{T}}\mathbf{v} - \mathbf{v}^{\mathrm{T}}\mathbf{W}\mathbf{h} \tag{1}$$

As in general Boltzmann machines, the probability distribution over hidden and visible vectors are defined as

$$p(\mathbf{v},\mathbf{h}) = \frac{1}{Z}e^{-E(\mathbf{v},\mathbf{h})}, \quad Z = \sum_{\mathbf{h}} e^{-E(\mathbf{v},\mathbf{h})} \tag{2}$$

The marginal probability of the visible vector is

$$p(\mathbf{v}) = \frac{1}{Z}\sum_{\mathbf{h}} e^{-E(\mathbf{v},\mathbf{h})} \tag{3}$$

Since there are no direct connections between two hidden units (Fig. 1), the hidden units conditioned on **v** are independent of each other. Similarly, the visible units conditioned on **h** are also independent of each other. The units of a binary hidden layer, conditioned on the visible layer, are independent Bernoulli random variables. The binary state $h_j$ of the hidden unit $j$ is set to 1 with probability

$$p(h_j = 1|\mathbf{v}) = \delta\left(\sum\nolimits_i w_{ij}v_i + b_j\right) \tag{4}$$

where $\delta(x) = 1/(1+\exp(-x))$ is the sigmoid function. If the visible units are binary, the visible units, conditioned on the hidden layer, are also independent Bernoulli random variables. In this case, the binary state $v_i$ of the visible unit $i$ is set to 1 with probability

$$p(v_i = 1|\mathbf{h}) = \delta\left(\sum\nolimits_j w_{ij}h_j + c_i\right) \tag{5}$$

If the visible units have real values, then the visible units, conditioned on the hidden layer, are independent Gaussian random variables defined as

$$p(v_i\,|\mathbf{h}) = g\left(\sum\nolimits_j w_{ij}h_j + c_i\,, 1\right) \tag{6}$$

where $g(.)$ represents the Gaussian distribution. Since the RBM is a generative model, suppose $\boldsymbol{\theta} = \{\mathbf{W}, \mathbf{b}, \mathbf{c}\}$ contains the parameters of a RBM. The parameters can be calculated by performing stochastic gradient descent on the log-likelihood of training samples. The probability that the network assigns to a sample $v^{(k)}$ ($k$ = 1…K) is given by summing over all possible hidden vectors as

$$\arg\min_{\boldsymbol{\theta}} -\sum_k \log\left(\sum_{\mathbf{h}} e^{-E(\mathbf{v}^{(k)},\mathbf{h}^{(k)})}\right) \tag{7}$$

By solving Eq. 7, one could calculate the parameters $\boldsymbol{\theta}$ offline, and use them to configure a RBM. And a DBN can be built by stacking multiple layers of RBMs trained in a layer-by-layer manner.

### *B. Adaptive RBM with Mixed Norm Regularization*

For efficient embedded implementations, we propose a sparsely weighted variant of the RBM, named as the Adaptive RBM (AdaRBM), which adds an extra regularization term in Eq. 7 to shrink the weights. Before introducing the AdaRBM algorithm, we first define the mixed norm of a matrix **W** as

$$\|\mathbf{W}\|_{\mathrm{M}} = \sum\nolimits_i \left(\sum\nolimits_j |w_{i,j}|^2\right)^{1/2} \tag{8}$$

where the two indices $i$ and $j$ are treated differently. It is easy to prove that the mixed norm is indeed a legitimate matrix norm, and it is different from the standard $L_1$ or $L_2$ matrix norms defined as follows

$$\|\mathbf{W}\|_{\mathrm{L}_1} = \sum_i\sum_j |w_{i,j}|, \|\mathbf{W}\|_{\mathrm{L}_2} = (\sum_i\sum_j w_{i,j}{}^2)^{1/2} \tag{9}$$

The mixed matrix norm defined in Eq. 8 adds the vector norms of all rows in a matrix, and minimizing the mixed norm could reduce the lengths of the matrix's rows. It is worth noting that the shrinking process doesn't apply evenly to all rows. Specifically, in the stochastic gradient descent process, shorter rows shrink faster than the rows with larger weights. And the weights in short rows tend to shrink to zero after finite iterations. Similarly, minimizing the mixed norm of a transposed matrix like $\mathbf{W}^{\mathrm{T}}$ could reduce the weights in short columns to zero.

In practice, the training algorithms of deep neural networks are designed to extract important features for classification. By reducing the mixed-norm of the two matrices **W** and $\mathbf{W}^{\mathrm{T}}$, the AdaRBM could shrink the weight parameters in short rows and columns to zero, therefore could be used to select important input and output features. Suppose the $i$th row of the weight matrix **W** is reduced to zero, then the output **h** will not be affected by the input feature $v_i$. On the other hand, suppose all the weight parameters $w_{ij}$ associated with $h_j$ is reduced to zero, then according to Eq. 4, given any evaluating samples，the probability $p(h_j = 1|\mathbf{v})$ will stay close to a constant $\delta(b_j)$, leading to negligible output feature $h_j$ for higher level of abstraction or classification.

As an unsupervised feature selection method, the AdaRBM assumes that not all features in **v** or **h** are required for higher level of abstraction, and the AdaRBM aims to select the input

TABLE I

| Training algorithm of the Adaptive RBM |
| --- |
| 1) Given $\langle v_i h_j \rangle_{\text{model}}$ represents a distribution defined by running a Gibbs chain, update using the contrastive divergence rule as $w_{ij} = w_{ij} + \epsilon(\langle v_i\, h_j\rangle_{\text{data}} - \langle v_i\, h_j\rangle_{\text{model}})$ $b_i = b_i + \epsilon(\langle h_j\rangle_{\text{data}} - \langle h_j\rangle_{\text{model}})$ $c_i = c_i + \epsilon(\langle v_i\rangle_{\text{data}} - \langle v_i\rangle_{\text{model}})$ where $\epsilon$ is a learning rate, and $\langle . \rangle_{\text{model}}$ is an expectation over thee reconstruction data, estimated using one iteration of Gibbs sampling; |
| 2) Update the parameters using the gradient of $R_s(\mathbf{W})$ as $w_{ij} = w_{ij} + \epsilon * \lambda\, (\gamma \frac{\mathrm{w}_{ij}}{\sqrt{\sum_i \mathrm{w}_{ij}}} + (1-\gamma)\frac{\mathrm{w}_{ij}}{\sqrt{\sum_j \mathrm{w}_{ij}}})$ |
| 3) Check the constraint and repeat the update process until it convergence. |

and output features by shrinking the following regularization term

$$R_s(\mathbf{W}) = \lambda(\gamma\|\mathbf{W}\|_{\mathrm{M}} + (1-\gamma)\|\mathbf{W}^{\mathrm{T}}\|_{\mathrm{M}}) \tag{10}$$

where $\lambda$ is the parameter to control the sparsity of the weight parameters, $\gamma$ controls the balance between the row sparsity and column sparsity. And the AdaRBM training algorithm attempts to shrink the regularization term of Eq. 10 by incorporating it in the optimization of the standard RBM (Eq. 7) as

$$\operatorname{argmin}_{\boldsymbol{\theta}} - \sum_k \log\left(\sum_{\mathbf{h}} e^{-E(\mathbf{v}^{(k)},\mathbf{h}^{(k)})}\right) + \lambda(\|\mathbf{W}\|_{\mathrm{M}} + (1-\gamma)\|\mathbf{W}^{\mathrm{T}}\|_{\mathrm{M}}) \tag{11}$$

### C. Training Adaptive RBM and Deep Adaptive Network

Generally, the objective function of Eq. 11 is the sum of a log-likelihood term and a regularization term. The derivatives of the log probability and the regularization term with respect to the parameters can be expressed as

$$\frac{\partial \log p(\mathbf{v})}{\partial w_{\mathrm{ij}}} = \langle v_i h_j\rangle_{\text{data}} - \langle v_i h_j\rangle_{\text{model}} + \lambda\frac{\partial R_s(\mathbf{W})}{\partial w_{\mathrm{ij}}} \tag{12}$$

$$\frac{\partial \log p(\mathbf{v})}{\partial b_j} = \langle h_j\rangle_{\text{data}} - \langle h_j\rangle_{\text{model}} \tag{13}$$

$$\frac{\partial \log p(\mathbf{v})}{\partial c_{\mathrm{i}}} = \langle v_i\rangle_{\text{data}} - \langle v_i\rangle_{\text{model}} \tag{14}$$

$$\frac{\partial R_s(\mathbf{W})}{\partial w_{\mathrm{ij}}} = \gamma\frac{\mathrm{w}_{ij}}{\sqrt{\sum_i \mathrm{w}_{ij}}} + (1-\gamma)\frac{\mathrm{w}_{ij}}{\sqrt{\sum_j \mathrm{w}_{ij}}} \tag{15}$$

where $\langle\cdot\rangle_p$ indicates the expectation of distribution $p$. Unfortunately, similar to the RBM training process, the above equations are not tractable, because the computations of above expectations are very difficult; however, one could use the contrastive divergence (CD) with Gibbs sampling to approximate the optimal parameters in an iterative way. On each iteration we apply the contrastive divergence update rule, followed by one step of gradient descent of the regularization term as in Table 1.

It is worth noting that, according to Eq. 15, the weight decay process shrinks the weight parameters unevenly. After a few hundreds of iterations, the weights in short rows and columns could be reduced to values close to zero, which leads to sparse weight parameters.

Multiple layers of the AdaRBMs can be stacked to compose a DAN which consists of multiple hidden layers (Fig. 1), with two adjacent layers forming an AdaRBM. Similar with the DBN training process, the DAN can be trained in a layer-by-layer style. The AdaRBM training algorithm can be repeated several times to learn a deep, hierarchical model. Specifically, one could train the bottom Adaptive RBM with CD on the training data. The corresponding parameters $\boldsymbol{\theta}$ will be frozen and the hidden unit values will be inferred. These inferred values then serve as the input data to train next higher layer in the network to model the hidden layer representations of the first-layer AdaRBM. This process can be repeated to yield a deep architecture that is an unsupervised model of the training distribution.

The Deep Adaptive Network, built with stacked Adaptive RBMs, is designed for efficient hardware implementation. By reducing the weight parameters using mixed-norm regularization, the DAN algorithm could build a multilayer feedforward neural network with sparse connections which can be robustly thresholded for efficient binary representation. Technically, the DAN calculates the sparse-binary weights in three steps, (1) the DAN calculates the real-value sparse parameters according to algorithm 1 (Table 1); (2) the weight parameters are thresholded by a small value $u$, and the weights with small absolute values are omitted; (3) the remained positive and negative weights are binarized as +1 and -1 respectively.

### D. Properties of the Deep Adaptive Network

The challenge of implementing a DBN using embedded system is memory and computation bottleneck. First, the weight parameters may exhaust Gigabytes of memory and the bandwidth between the processing units and the memory. Secondly, the multiplications between the weight parameters and the input features are the most time-consuming in a feedforward neural network, which causes the bottleneck in computation. Thirdly, a typical hardware high-precision multiplier usually consists of hundreds of logic gates, which made it a challenge of implementing large deep neural networks in hardware. The DAN addresses these challenge by learning an efficient hardware-oriented network architecture featuring two properties:

1) The neural connections are sparse. The DAN adaptively reduce the weight parameters associated with negligible neural connections to zero, which can be omitted for efficient embedded hardware implementation.
2) The neural connections can be represented as binary integers. Since the sparse weights of the DAN are well separated (Fig. 2), therefore the weights can be can be robustly thresholded, and each neural connection can be represented using a single-bit. Single-bit binary weight parameters could save a major proportion of memory and relieve the bandwidth bottleneck. Furthermore, feedforward neural networks, single-bit representation allows one to replace the complicated high-precision multipliers with fast, area-efficient computation units, which relieved the computation bottleneck for digital implementations.

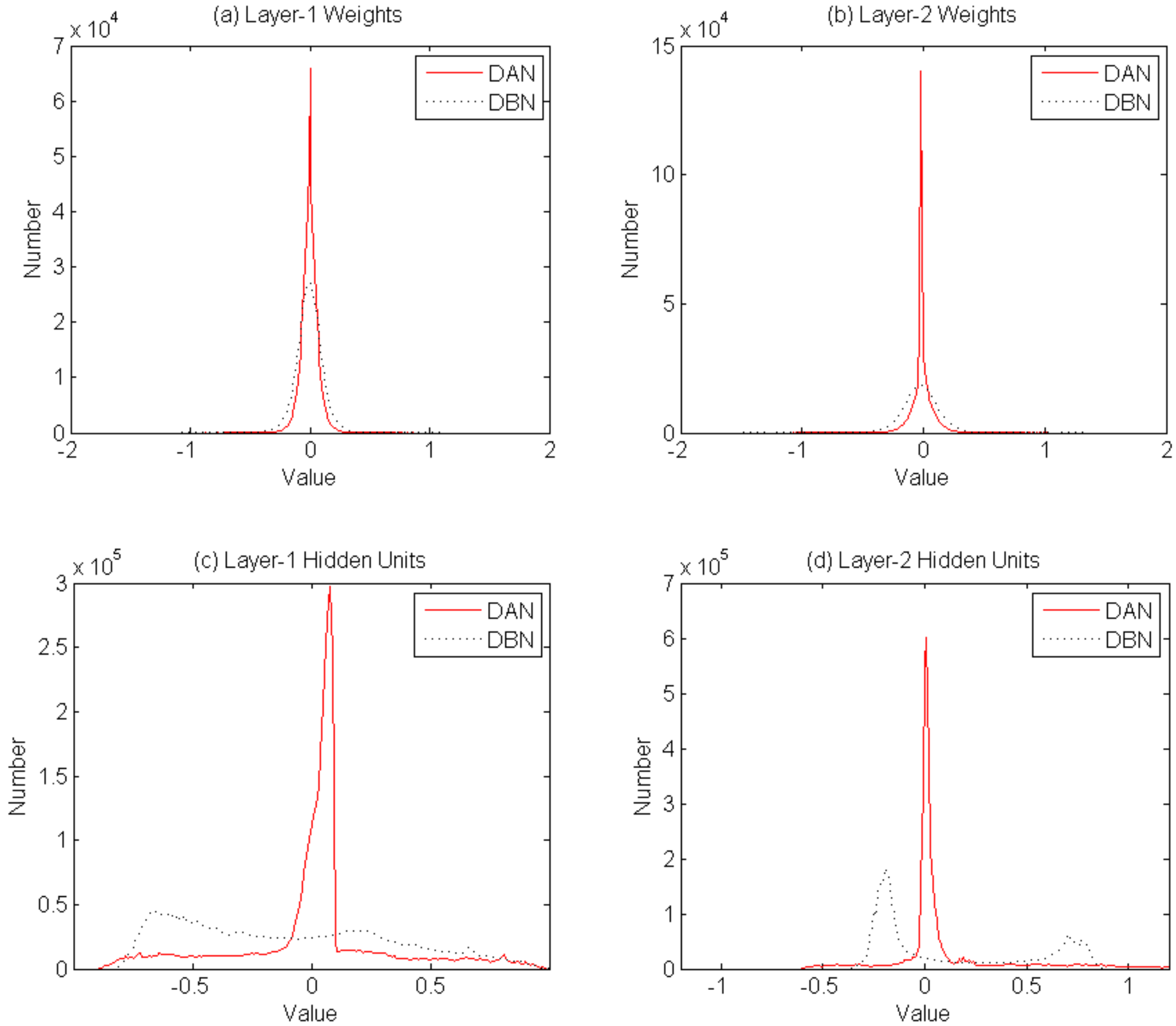

Fig. 3 Empirical distributions of the weight parameters and the hidden units of a two-layer DBN and DAN: Diagram (a-b) show the weight distribution, diagram (c-d) show the distribution of hidden units. The models are trained by 10000 handwritten images of randomly selected from the MNIST benchmark.

Figure 3 illustrates the DAN weight parameters and features. We implemented two two-layer deep neural networks of the DBN and the DAN in an experiment to select the features from the MNIST dataset. Both networks had the same width of $784 \times 800 \times 800$. Since the DAN adopts the mixed-norm regularization in the training process, most DAN weights were very close to zero (Fig. 3 a-b).

Similar to previous sparse variants of the DBN, the DAN could be used to extract sparse features. Specifically, during the training phase, the AdaRBM uses the probability of $h_j$=1 as the $j$th output feature to higher levels. Since the weights in short columns of the weight matrix are reduced to zero, the output features associated with zero columns become close to a constant independent from the input features. In our experiment, by subtracting $\delta(\mathbf{b})$, the hidden units of the stacked Adaptive RBMs became notably sparse (Fig. 3.c-d).

In summary, the DAN attempts to select sparse features of input samples by exploiting the sparsity of the weight parameters in a deep neural network. As shown in Fig. 3 b, the weight parameters become sparser in the second layer than the first layer. It seems that the sparse features tend to improve the sparsity of weight parameters in higher level AdaRBMs.

Through, motivated by different reasons, the $L_1$ norm regularization approach could also be used to calculate sparse weight parameters during the DBN training process [28]; however, our experiments over the MNIST dataset show that, the features extracted by the DBN with $L_1$-norm regularization has notably lower classification accuracy than the DAN; Moreover, $L_1$-norm weight-decay approach is not robust for single-bit representation. Specifically, applying thresholding and binarization to the weight parameters calculated by DBN with $L_1$-norm regularization results in a major drop of classification accuracy, from over 90% to less than 50%. On the other hand, the DAN algorithm is more robust with binary weight representation, which achieves 94% classification accuracy over the MNIST dataset.

## IV. EXPERIMENTS

We evaluated the DAN using the MNIST dataset for the application of recognizing handwritten digits.

### A. Experiment Setting and Measurements

As illustrated in Fig. 1, two networks of the DAN and the DBN were built to extract features from images of handwritten digits in the MNIST dataset. Both the DAN and the DBN had one layer of visible units and two layers of hidden units, which comprised two layers of stacked RBMs. The first-layer RBMs of both the DAN and the DBN had 784 visible units and 800 hidden units, and the second-layer RBMs had 800 visible and 800 hidden units. The MNIST handwritten digits were images with 784 pixels. We randomly selected 10000 images from the MNIST dataset to train the DAN and the DBN simultaneously.

In the experiments, we implemented the sparse and binary connected DAN in three steps. In each step, the feedforward neural network became more efficient. To distinguish these networks, we give the feedforward neural networks with different levels of precision respective names (Table 2) as

1) DAN: a real-valued Deep Adaptive Network learned according to Algorithm 1;
2) $DAN_s$: a sparse DAN whose weight parameters are thresholded to be zeros, and the remained weights have real values.
3) $DAN_b$: a sparse-connected binary-weighted DAN, which is calculated by substituting the positive and negative weights of the $DAN_s$ for +1 and -1 respectively.
4) $DAN_B$: a sparse-connected binary-weighted DAN whose hidden units of each AdaRBM are binarized using a threshold of 0.5.

Fig. 4 illustrates the $DAN_b$ parameters and hidden units learned from the images of the MNIST dataset. The weight parameters are shown in subgraph (b), where the weight matrices of the first-layer RBM and the second-layer RBM are sparse and binary. A third layer of full-connected neural network is connected with the DAN for classification. The weight matrix (800-10) of the classifier of classic neural network are real-valued. An example of input image and the hidden units of the first layer and second layer AdaRBMs are shown in subgraph (a). It is worth noting that the sparse units in the AdaRBM are close $\delta(\mathbf{b})$ rather than zero in subgraph (a).

TABLE II
COMPARISON OF DIFFERENT DEEP NEURAL NETWORKS[1]

| Methods | Weight precision (bit) | Feature precision (bit) | Weight Memory (Kbyte) | Classification accuracy (%) |
|---|---|---|---|---|
| DBN | 32 | 32 | 4950 | 97.3 |
| DAN | 32 | 32 | 4950 | 97.4 |
| $DAN_s$ [2] | 32 | 32 | 1238 | 97.2 |
| $DAN_b$ [3] | 1 | 32 | 30 | 94.0 |
| $DAN_B$ [3] | 1 | 1 | 30 | 93.3 |

1. All the deep neural networks (784-800-800) are built with two stacked layers used to extract features from the MNIST dataset.
2. The weight memory and classification accuracy are calculated with 25% important connections reserved.
3. The weight memory and classification accuracy are calculated with 20% important connections reserved.

To evaluate the sparsity of connections, we use a simple measurement σ which is the ratio of reserved weights controlled by a threshold value $u$ as

$$\sigma = 1 - \frac{\mathrm{Nth}(u)}{\text{Total number of weights}} \times 100\%$$

where $\mathrm{Nth}(u)$ indicates the number of weights whose absolute values are smaller than the threshold $u$. Specifically, the sparsity measurement $\sigma \in [0,1]$ reaches 0 if $u = \max(|w_{ij}|)$. The sparsity measurement is an important trade-off between efficiency and classification accuracy, and a typical range of σ for the DAN weight parameters is between 5%-25% for the MNIST dataset.

Different classifiers could be applied for recognizing the handwritten digits. Since the classic neural networks are naturally parallel, most hardware accelerators of the DBN adopts the neural network classifier for pattern recognition [9-13]. In our experiment, we classified the features extracted by the DAN and DBN using a fully-connected 800-by-10 neural network. The classification accuracy of the neural network over 10000 randomly selected test images were calculated to evaluate the features extracted by the DAN.

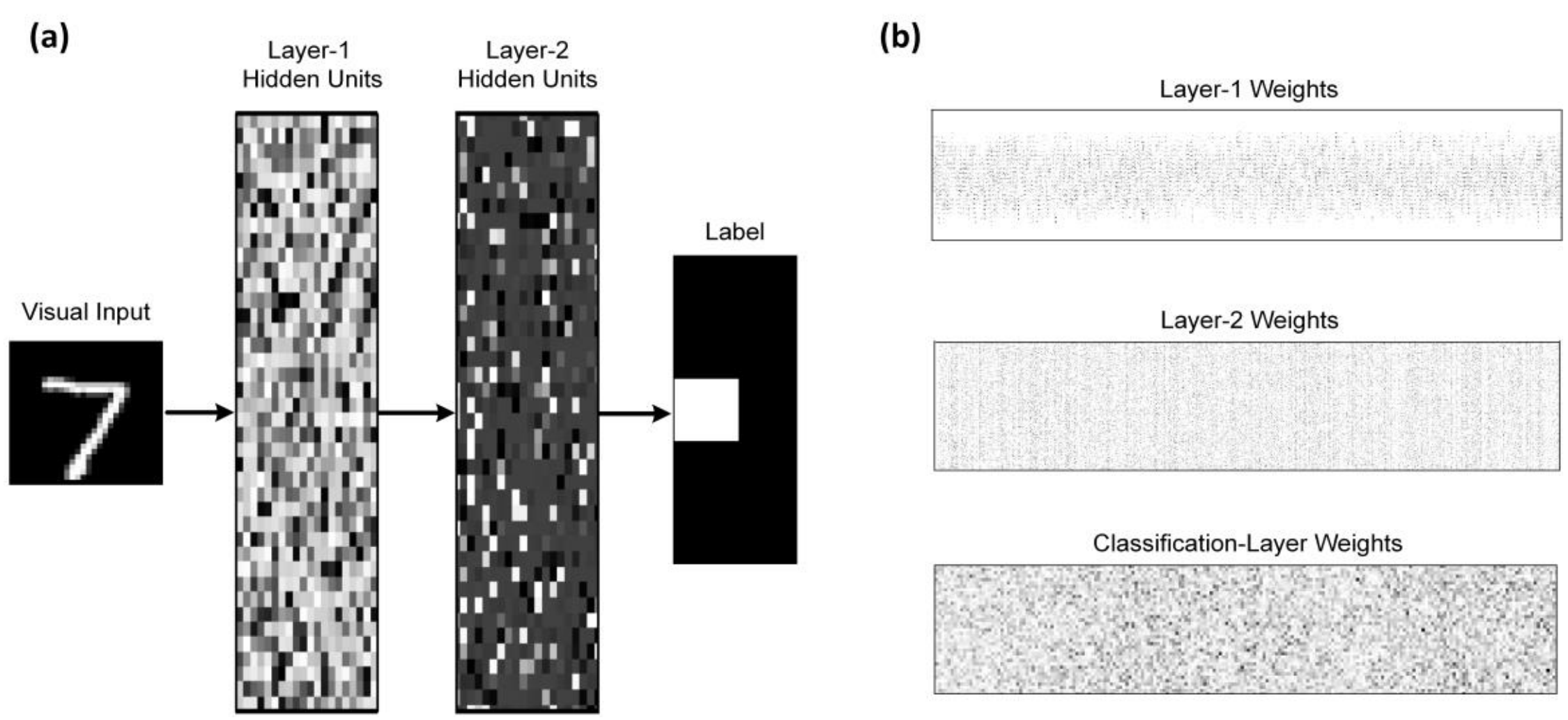

Fig. 4. The sparsely-connected binary-weighted Deep Adaptive Network (DAN) learned from images of handwritten digits in the MNIST dataset. A third layer of full-connected neural network is connected with the DAN for classification.

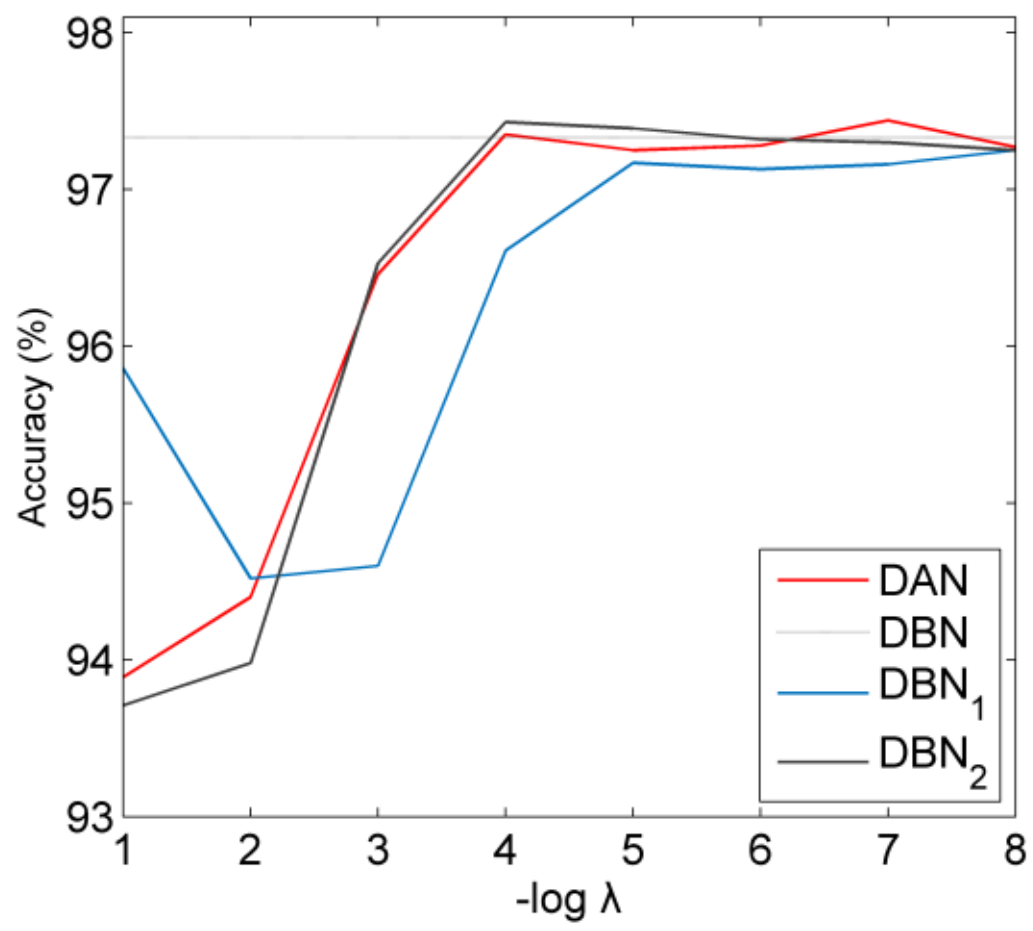

Figure 5. The classification accuracy of the DBNs and the DAN changes as the regularization parameter $\lambda$ changed from $10^{-1}$ to $10^{-8}$ ($\gamma$= 0.5).

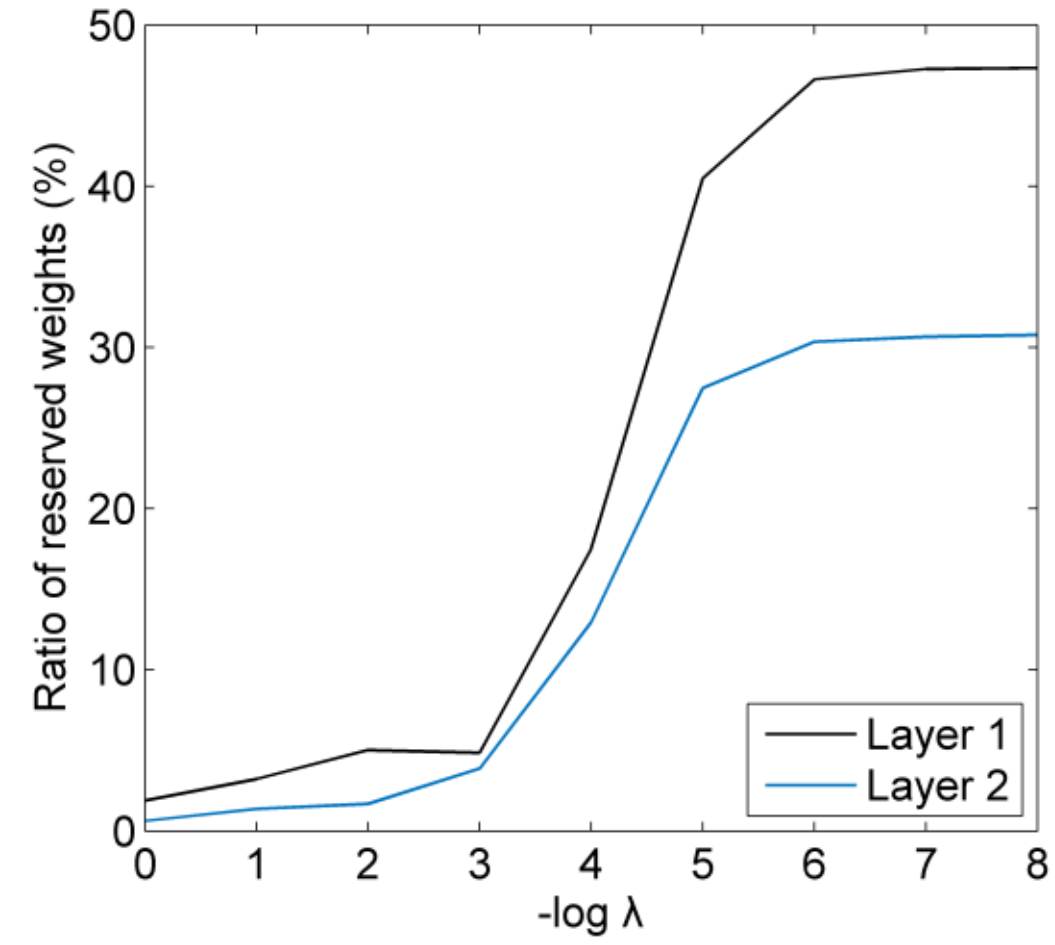

Figure 6. The ratio of reserved weights (σ%) for each layer of the DAN is controlled by the parameter $\lambda$ ($u$ = 0.1, $\gamma$= 0.5). It is worth noting that the second layer is sparser than the first layer.

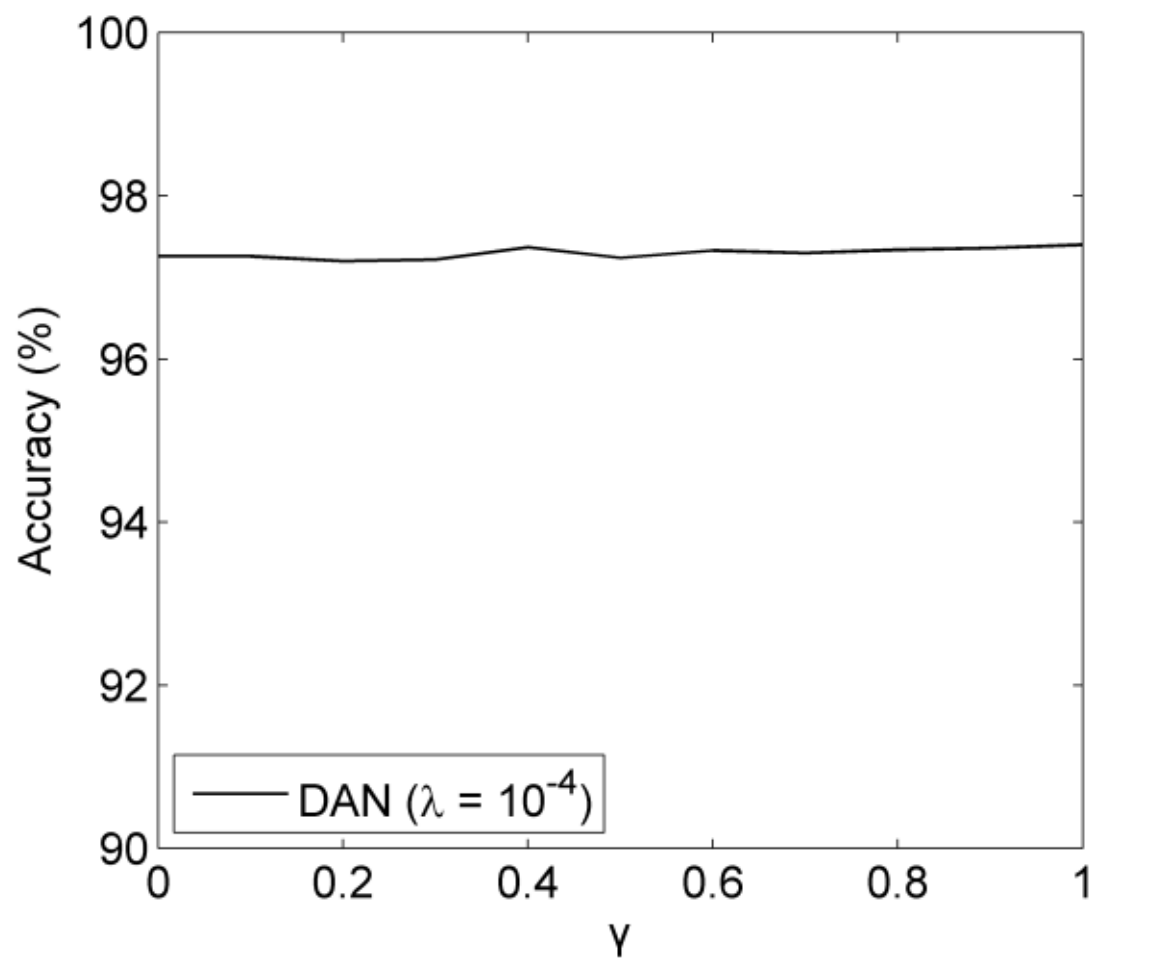

Figure 7. The classification accuracy of the DAN is robust as the regularization parameter $\gamma$ changes.

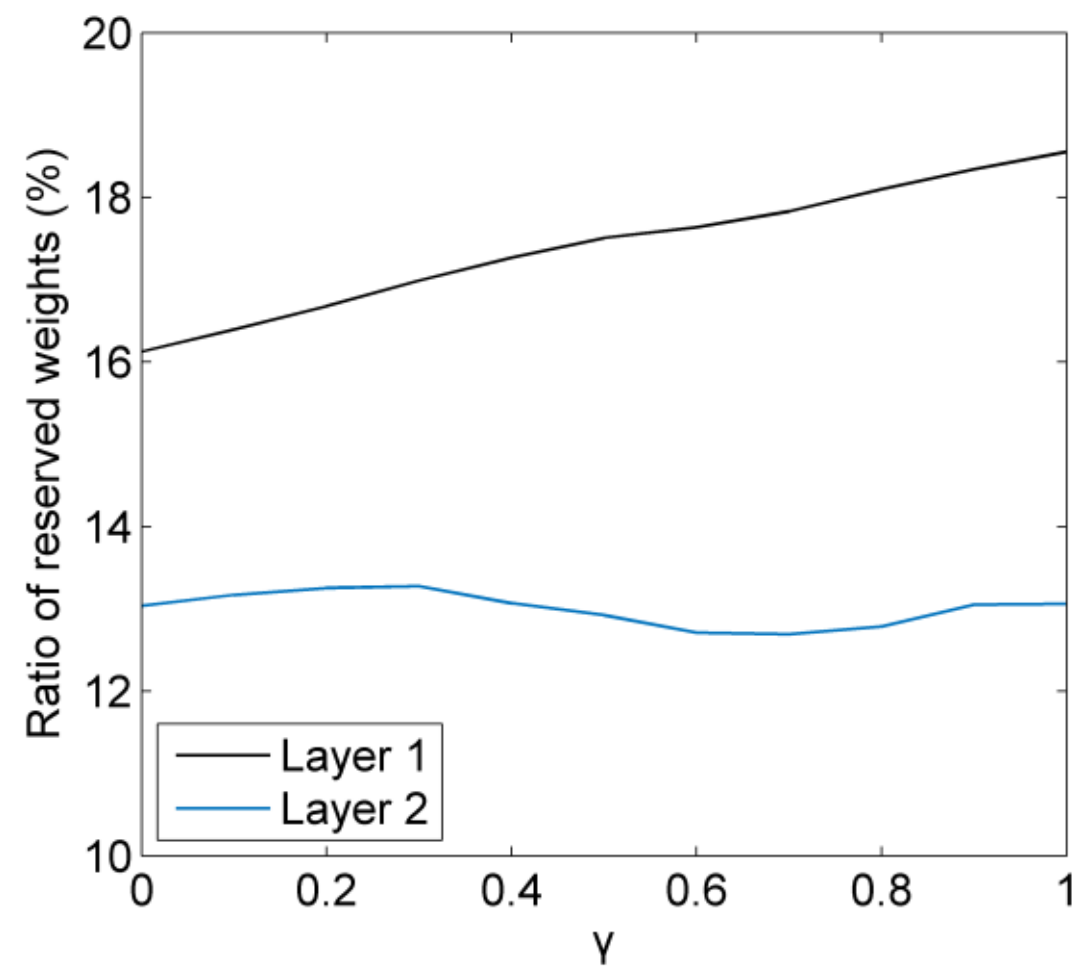

Figure 8. The ratio of reserved weights for the DAN is moderately affected by the parameter γ for the MNIST dataset ($\lambda$=$10^{-4}$).

### *B. Choosing Parameter λ*

The DAN training algorithm has two parameters, i.e. $\lambda$ and $\gamma$. Parameter $\lambda$ controls the sparsity of the weight parameters. In the first experiment, we changed the parameter λ to examine its relation to classification accuracy. Four types of deep neural networks were compared, i.e. the DAN, the original DBN, the DBN with $L_1$-norm weight-decay ($DBN_1$) and the DBN with $L_2$-norm weight-decay ($DBN_2$). We changed the parameter $\lambda$ from $10^{-1}$ to $10^{-8}$. As shown in Fig. 5, the DAN, DBN and $DBN_2$ had similar classification accuracy about 97% when λ is smaller than $10^{-4}$; however, the $DBN_1$ has notably lower accuracy (96%) than the other algorithms. Experiment shows that the DAN and DBN show better pattern recognition results than the original DBN when the parameter λ is equal to or smaller than $10^{-4}$. The reason for that, as suggested by Hinton, might be because adding a small regularization term could reduce overfitting and improve classification accuracy [28].

On the other hand, the parameter λ could affect the efficiency of hardware implementation. As indicated by Eq. 11, the parameter λ controls the sparsity of the weight parameters in a Deep Adaptive Network. We examined the relation between the ratio of the reserved weights (σ) and the value of λ. We set a typical threshold value $u$=0.1 and γ= 0.5, and changed λ from $10^{-1}$ to $10^{-8}$. Fig. 6 shows that the ratio of reserved weights drops significantly from about 50% to less than 5% for the MINST dataset. It is worth noting that the second layer of the DAN is over 15% sparser than the first layer. Combining the results of classification accuracy and weight sparsity, the recommended range of parameter λ should be $10^{-2}$ to $10^{-4}$ for the MNIST dataset.

### *C. Choosing Parameter γ*

To increase flexibility, a second parameter $\gamma$ is added in the DAN formulation (Eq. 10). We designed a second group of experiments to examine the parameter $\gamma$. A series of DANs with different $\gamma \in [0,1]$ were trained to observe how the parameter $\gamma$ affected classification accuracy and the sparsity of weights. The parameter $\gamma$ was changed from 0 to 1 with step size of 0.1. As shown in Fig. 7 the classification accuracy of the DAN is robust with varied $\gamma$.

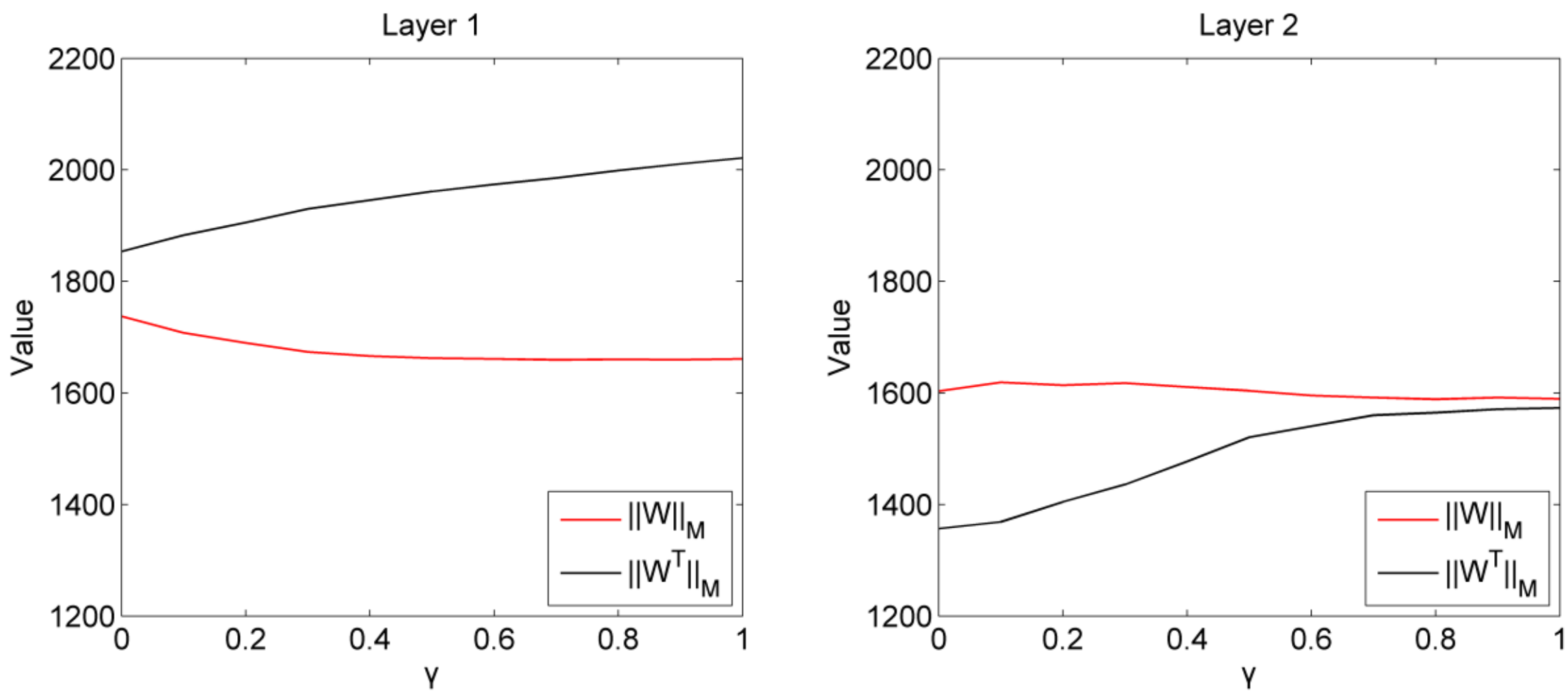

Figure 9. The regularization term $||\mathbf{W}||_M$ and $||\mathbf{W}^T||_M$ changed as $\gamma$ changed, where the $||\mathbf{W}||_M$ represents the sum of the length for each row of $\mathbf{W}$, $||\mathbf{W}^T||_M$ represents the sum of length for each column of $\mathbf{W}$.

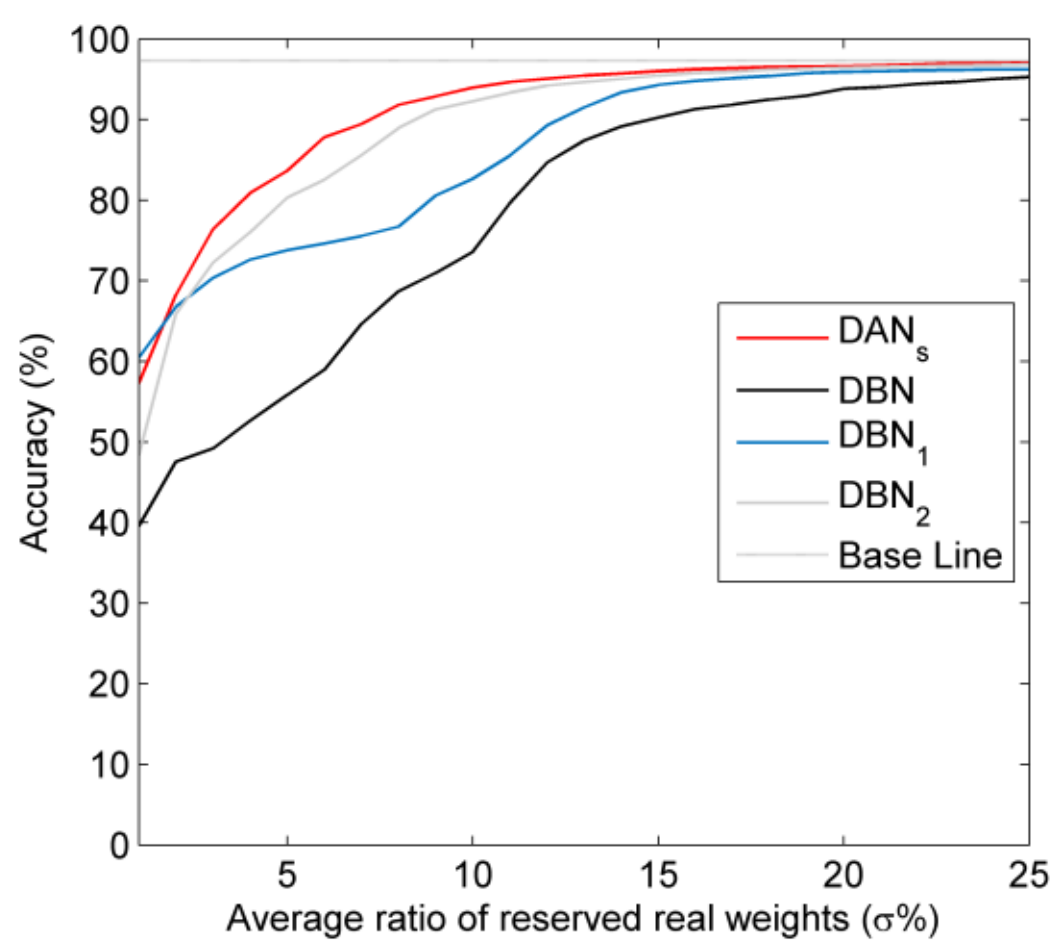

Figure 10. Classification accuracy of the thresholded sparse DAN ($DAN_s$) and the DBNs with $L_1$ norm regularization ($DBN_1$) and $L_2$ norm regularization ($DBN_2$). The weights of the $DAN_s$ and DBNs are thresholded, and only a proportion of (σ%) weights with the largest absolute values are reserved. The base line is the classification accuracy of the original DBN with high-precision parameters.

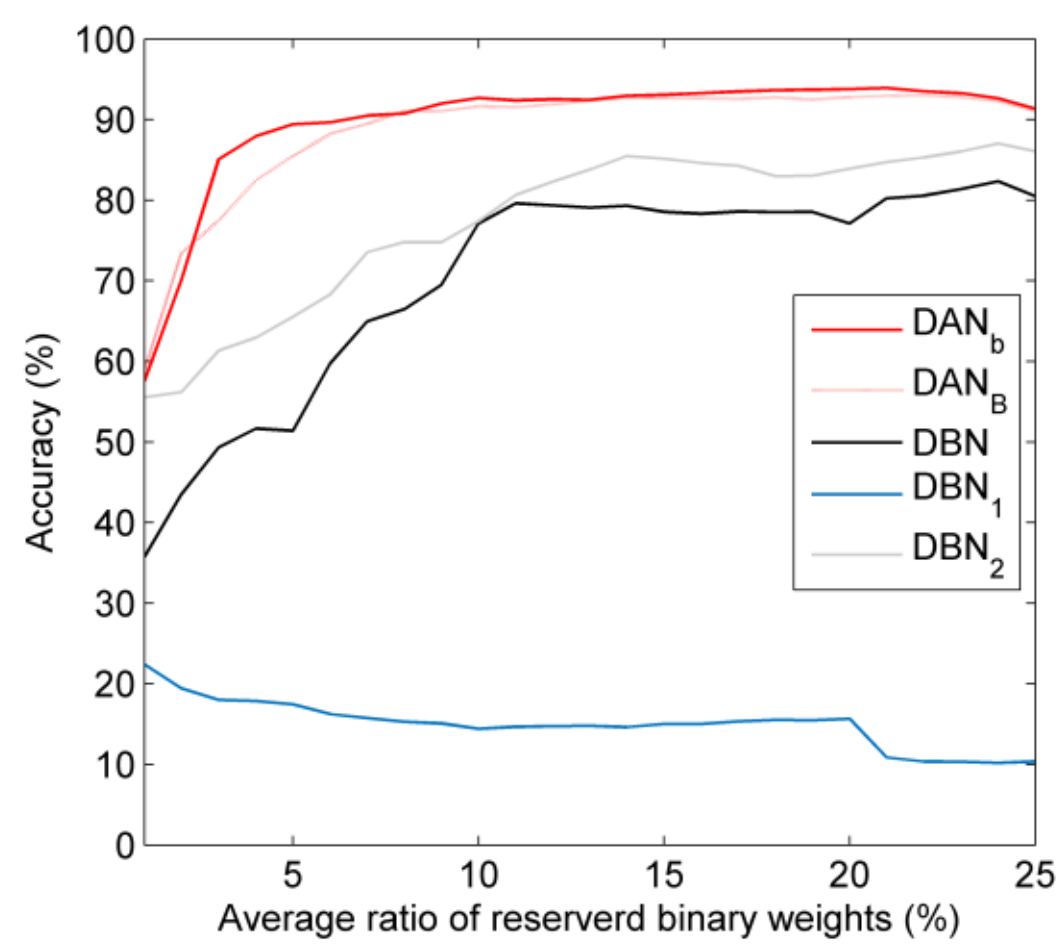

Figure 11. Classification accuracy of the sparse-binary-weighted DAN ($DAN_b$) and the DBNs with $L_1$ ($DBN_1$) and $L_2$ norm ($DBN_2$) regularization terms. The weights of the DAN and DBNs are thresholded with the σ% significant weights reserved, then the reserved are represented as +1 and -1 according to the signs. The base line is the classification accuracy of the original DBN without threshold operation.

As for sparsity, we examined the relation between the ratio of the reserved weights and the value of γ. We set a typical value of $\lambda=10^{-4}$ and changed γ from 0 to 1.0 with 0.1 step size. Fig. 8 shows that for the first layer of AdaRBM, the preferable value of γ is 0, which achieves 16% sparsity. For the second layer of AdaRBM, γ = 0.7 is the preferable value which leads to about 13% sparsity for the weight parameters.

Since the parameter $\gamma$ controls the trade-off between the row sparsity and column sparsity of the weight matrix (Eq. 10), we examined the weights of the DANs with different $\gamma$. Fig. 9 compares the row-wise mixed norm $||\mathbf{W}||_M$ and the column-wise mixed norm $||\mathbf{W}^T||_M$, and one has two observations. First, $||\mathbf{W}||_M$ increases as $\gamma$ increases, whereas $||\mathbf{W}^T||_M$ drops as $\gamma$ increases. This observation is coherent with the optimization formulation of Eq. 11. Second, the weight matrix of the second layer AdaRBM has smaller mixed norm than the first-layer AdaRBM. This observation indicates that the second-layer weights may be sparser than the first-layer weights, which is coherent with the results of Fig. 6.

### *D. Deep Adaptive Network with Thresholded Sparse Weights*

Since the weight parameters of the DAN are sparse and separated into roughly three groups, one can robustly threshold the weights using a small value *u*. The proportion of reserved weights σ decreases when *u* increases. By omitting the weights smaller than *u*, the DAN can be efficient implemented with almost no influence on pattern recognition performance.

Fig. 10 compares the classification accuracy of a classic neural network proceeded by different variants of the Deep Belief Networks. The base line shows the classification accuracy (97.3%) of the original DBN with 32-bit high-precision weights. The accuracy for the DAN with thresholded sparse weights ($DAN_s$) increases from 58.6% to 97.3% when the average proportion of reserved weights for the first layer and second layer increases from 1% to 25%. Fig. 10 compares the $DAN_s$ with the original DBN, the DBN with $L_1$-norm weight decay ($DBN_1$) and the DBN with $L_2$-norm weight decay ($DBN_2$). Since the images of handwritten digits are relatively sparse (Fig. 4), all the algorithms shows above 90% classification accuracy when the ratio of the reserved rates exceeds 20%; however, given lower ratio of reserved weights, e.g. σ = 10%, the $DAN_s$ (92.6%) shows notably higher accuracy than the original DBN (70.9%), the $DBN_1$ (80.5%) and the $DBN_2$ (89.8%).

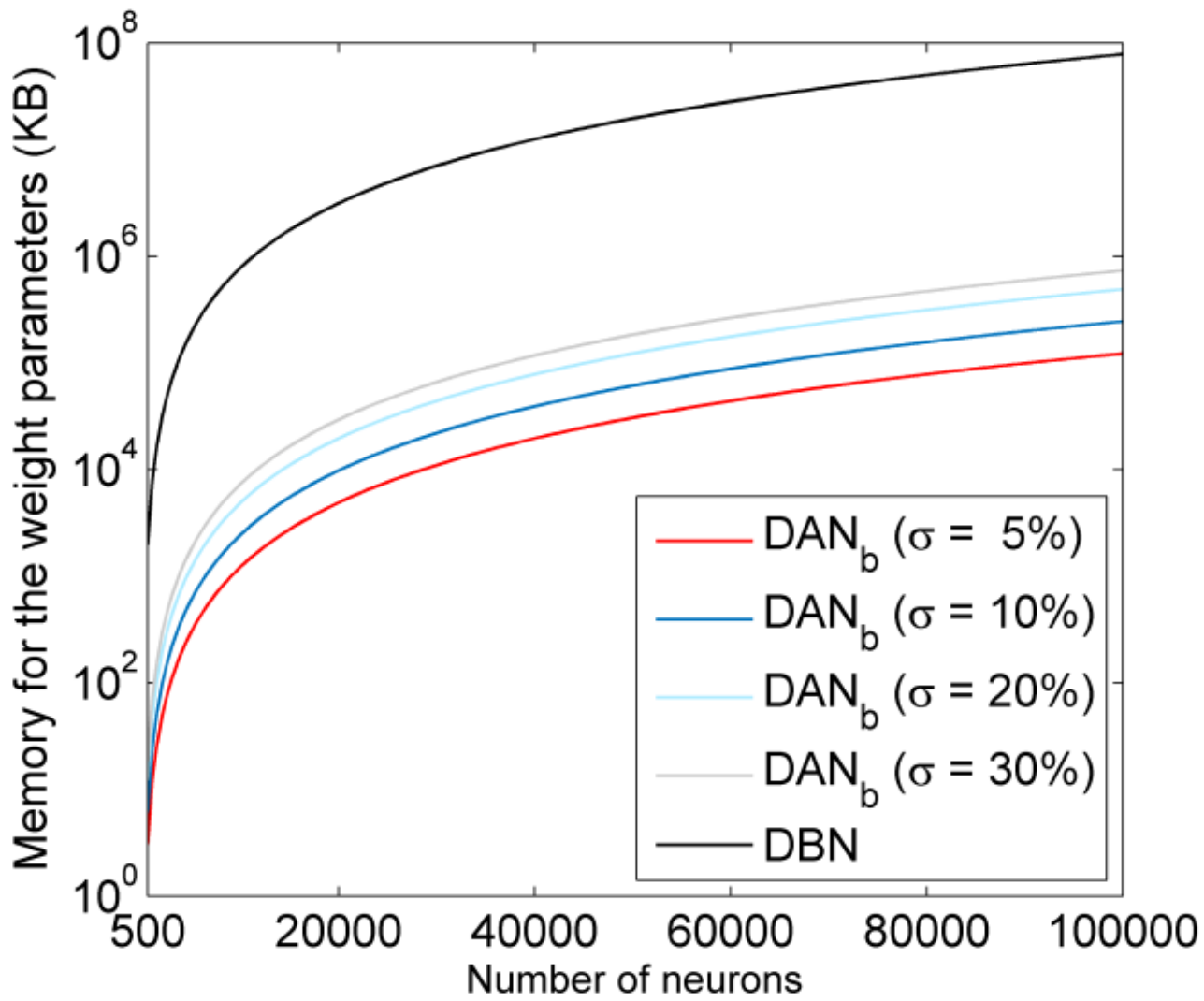

Figure 12. Memory required to store the weight parameters for the DBN and the DAN with sparse and binary weights ($DAN_b$). The weight memory of the DAN increases as the ratio of reserved rates σ increases.

### *E. Deep Adaptive Network with Sparse and Binary Weights*

Since the DAN weight parameters are naturally separated, the significant positive and negative weights of the $DAN_s$ could be represented by one bit integers (+1 or -1) without significant influence on classification accuracy. Fig. 11 compares the robustness of binarization operation for the thresholded $DAN_s$ and the DBNs with $L_1$ and $L_2$ norm regularization. The $DAN_s$ with thresholded binary weights is represented as $DAN_b$ and the DBNs' weights were also thresholded and binarized.

The DAN is sparse and robust with binarization operation. With only 10% single-bit weights reserved, the classification accuracy of the $DAN_b$ is 92.0 %, whereas, the classification accuracy of the DBN (70.2 %), $DBN_1$ (15.3%) and $DBN_2$ (75.1%) drops by 20% to 80% over the MNIST dataset. The $DAN_b$ reaches the highest accuracy of 94.2% with 22% binary weights reserved, which is only 3% less than the original DBN with high-precision all-reserved weights.

By adopting sparse and binary weights, the $DAN_b$ is over 99% more efficient than the original DBN in terms of memory consumption. Fig. 12 shows the memory used by the $DAN_b$ and the original DBN. The memory complexity of deep neural networks increases when the number of neurons in the network increases. The original DBNs generally implemented using 32-bit high-precision representation in previous research [28]; however, with sparse and binary weights, the $DAN_b$ generally improve the memory efficiency by about two (σ = 30%) to three (σ = 5%) orders of magnitude.

The $DAN_b$ also improves computational efficiency by replacing complicated floating-point operations with simple fix-point operations. Specifically, the fundamental computations of a feedforward neural network are the huge number of multiplications between the weight parameters and the features. To achieve the highest efficiency for digital implementations, the DAN could use binary features as well as binary weights. Fig. 11 shows the influence of adopting binary features on classification accuracy. The $DAN_B$ is a Deep Adaptive Network with binary and sparse connections, and the features in the $DAN_B$ are thresholded and binarized using probability of 0.5. As shown in Fig. 11, the classification accuracy of the DAN is robust for adopting binary weights and features. Specifically, with only 10% binary weights reserved, the classification accuracy of the $DAN_B$ reaches 91.3% over MNIST dataset. It is worth noting that, by adopting binary weights and features, one could replace the complicated the complicated floating-point multipliers, which usually consists of hundreds of logic gates, with merely one logic gate, leading to three to four orders of magnitude of higher time and area efficiency for digital hardware implementations.

## V. Discussion and Conclusion

To bridge the gap between hardware and algorithmic researches, this paper demonstrated a deep learning algorithm specifically designed for digital hardware modeling. The proposed Deep Adaptive Network attempt to learn a deep neural network with sparse discrete connections which can be further quantized using single-bit integers. The impact of such a method on specialized hardware implementations of deep networks could be major, by removing the need of memory to store the huge number of weight parameters by over 99%, and replace the complex floating-point multipliers, typically implemented using hundreds of logic gates by one fix-point accumulator or even one simple logic gate for each neural connection, which could potentially reduce the footprints of the computing units by over 99%. Furthermore, with simpler computing requirement and significantly less parameters, the latency of feedforward neural computing and data transaction can be significantly reduced, leading to faster, larger, and more efficient hardware implementation of neural networks.